%% file: main.tex
\documentclass[10pt,twocolumn,letterpaper]{article}

\usepackage{cvpr}  
\usepackage{pifont}
\newcommand{\cmark}{\ding{51}} 
\newcommand{\xmark}{\ding{55}} 
\usepackage{graphicx}
\usepackage{tabularx}
\usepackage{booktabs}
\usepackage{algorithm}
\usepackage{amsmath}
\usepackage{algpseudocode}

\definecolor{cvprblue}{rgb}{0.21,0.49,0.74}
\usepackage[pagebackref,breaklinks,colorlinks,allcolors=cvprblue]{hyperref}

\def\paperID{30} 
\def\confName{CVPR}
\def\confYear{2026}

\title{CARL-CXR: Continual Adapter-Based Routing for Task-Unknown Chest Radiograph Classification}

\author{%
  \begin{tabular}[t]{cccc}
    Muthu Subash Kavitha$^*$ & Anas Zafar$^*$ & Amgad Muneer & Jia Wu$^\dagger$
  \end{tabular}\\
  Department of Imaging Physics, The University of Texas MD Anderson Cancer Center, Houston, TX, USA\\
  $^*$Equal contribution \quad $^\dagger$Corresponding author\\
  {\tt\small email@mdanderson.org}
}

\begin{document}
\maketitle
\input{sec/0_abstract}    
\input{sec/1_intro}

{
    \small
    \bibliographystyle{ieeenat_fullname}
    \bibliography{main}
}


\end{document}

%% file: sec/0_abstract.tex
\begin{abstract}
Clinical deployment of chest radiograph classifiers requires models 
that can be updated as new datasets become available without retraining 
on previously observed data or degrading validated performance. We 
study a task-incremental continual learning setting for chest 
radiograph classification under task-unknown inference, where 
heterogeneous chest X-ray datasets arrive sequentially and task 
identity is unavailable at deployment time. We propose CARL-CXR, a 
continual adapter-based routing framework that maintains a fixed 
high-capacity backbone while incrementally introducing lightweight 
task-specific adapters and classifier heads. A latent task selector 
operates on adapter-conditioned features to dynamically route each 
input to the most relevant task pathway, leveraging compact task 
prototypes and feature-level experience replay to preserve task 
identity across sequential updates without storing raw images. 
Experiments on MIMIC-CXR and CheXpert two large-scale datasets 
with distinct patient populations, imaging devices, and annotation 
pipelines demonstrate that CARL-CXR achieves minimal catastrophic 
forgetting (0.012 AUROC drop), representing a $6\times$ and 
$11\times$ reduction over established continual learning baselines 
LwF and EWC respectively, while maintaining competitive diagnostic 
performance (AUROC 0.74). Under task-unknown deployment, CARL-CXR 
outperforms joint training by 12.5 points in routing accuracy 
(75.0\% vs.\ 62.5\%): unlike LwF and EWC, which require explicit 
task identifiers at inference and provide no routing mechanism, 
joint training supports task-unknown routing but degrades 
substantially due to reduced task-specific feature separation, 
while CARL-CXR preserves reliable routing through its 
isolate-then-freeze adapter strategy. Feature-level experience 
replay proves essential for routing stability, improving balanced 
routing accuracy from 13.1\% to 65.3\% over a prototype-only 
baseline. The combined adapters and selector introduce only 
2.3\,MB of additional parameters approximately $1250\times$ 
fewer than full backbone fine-tuning providing a practical and 
parameter-efficient foundation for sequential clinical deployment.
\end{abstract}

%% file: sec/1_intro.tex
\section{Introduction}
\label{sec:intro}

Deep learning for chest radiograph classification achieves strong multi-label performance for common findings, especially with large backbones trained on curated datasets \cite{rajpurkar2017chexnet,irvin2019chexpert,johnson2019mimiccxr,zafar2026beyond}. This standard practice trains on a fixed distribution and retrains as new data arrive. However, it is less suitable for clinical deployment. 

A practical clinical system needs sequential updates across years while preserving earlier diagnostic knowledge. The requirement is threefold: new datasets should be added incrementally, earlier performance should remain stable without continuous access to historical images, and the update cost should remain small. Continual learning formalizes these requirements through the 
stability-plasticity framework~\cite{parisi2019continual}, 
which seeks to balance retention of prior knowledge with adaptation 
to new data. Several established continual learning approaches, however, 
remain difficult to adapt to large-scale medical backbones. Full network 
retraining increases cross-task interference and imposes significant 
storage burden, as all prior data must remain accessible~\cite{rebuffi2017icarl}. 
In medical imaging, data access constraints and computational cost 
are primary determinants of method choice.

Foundation radiograph models have demonstrated improved transferability and robustness under distribution shift \cite{wang2022medclip,tiu2022chexzero,zhang2020convirt, azizi2021bigselfsupervised}. However, adaptation to new clinical sources still typically relies on full layers fine-tuning or joint multi-dataset training \cite{kulkarni2025flcxr,chambon2022roentgen, ma2025fully}. Systematic evaluation of performance retention across repeated updates is essential for reliable clinical decision support \cite{muneer2025foundation}. This work introduces a continual chest radiograph classifier that supports sequential dataset ingestion with retention control. A Swin Transformer encoder is used as a frozen backbone to provide stable hierarchical features across time \cite{liu2021swin}. Each dataset is assigned a lightweight adapter and a task-specific head. 
Recent work in medical vision–language learning also supports frozen-backbone designs with lightweight modules to reduce training cost while preserving prior knowledge \cite{qin2024freeze_backbones}. 
A latent task selector is trained to infer the most appropriate task context from adapted feature representations, guided by compact task prototypes. Selector stability is maintained through feature-level experience replay, which stores a bounded set of adapted feature vectors across tasks. This mechanism preserves prior-task evidence for routing across updates and scales naturally as new datasets are added. The resulting system supports continual extension to additional datasets over time while maintaining stable performance and reliable task routing under label-free deployment. 
Our contributions are as follows:

\begin{figure*}[ht]
  \centering
  \includegraphics[width=\textwidth]{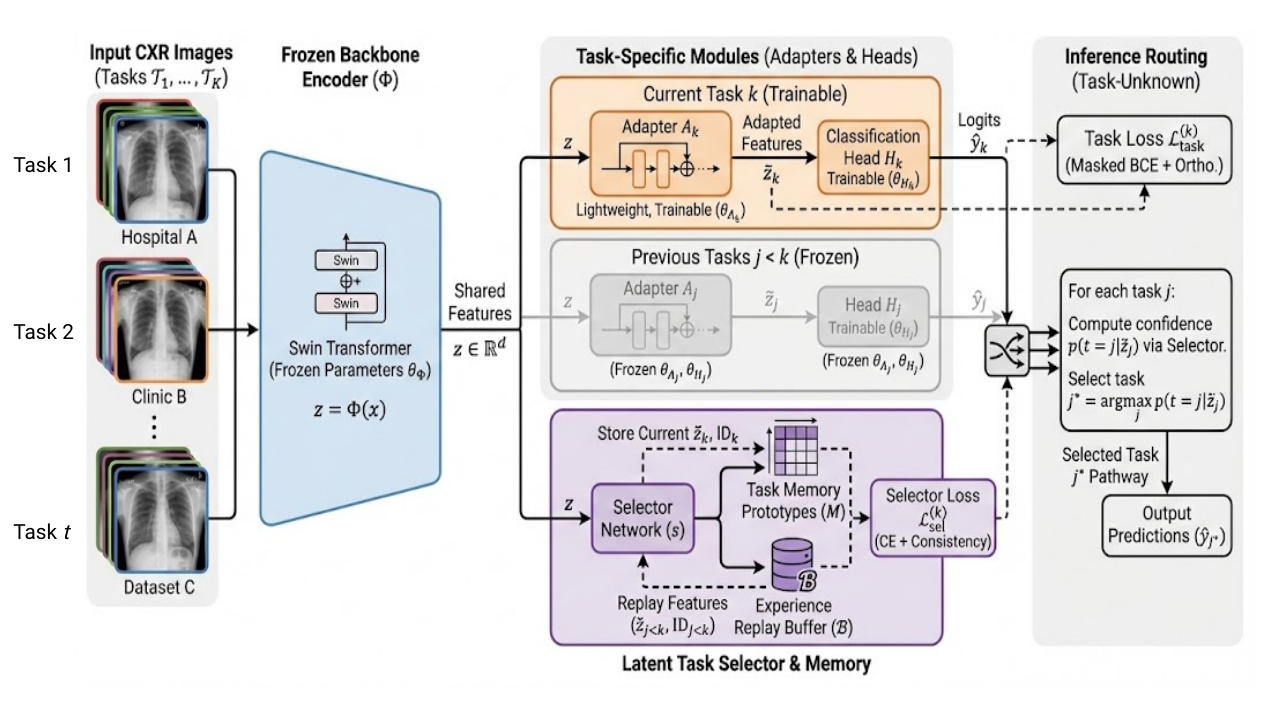}
  \caption{CARL-CXR framework: Architectural overview of continual chest radiograph learning with a frozen backbone, incremental task-specific adapters, and routing-based task selection during inference.}
  \label{fig:cl_framework}
\end{figure*}

\begin{itemize}
  \item We introduce the first task-incremental continual learning formulation for chest radiograph classification under task-unknown inference, where heterogeneous datasets arrive sequentially and task identifiers are unavailable at deployment. This setting reflects realistic clinical conditions and establishes a practical evaluation framework for continual radiograph learning, for which standardized protocols are currently lacking.

\item We propose CARL-CXR, a continual adapter-based routing framework that maintains a fixed high-capacity backbone while incrementally introducing lightweight task-specific adapters and classifier heads. A latent task selector operates on adapter-conditioned features to dynamically infer the most appropriate task pathway, while compact task prototypes and feature-level experience replay stabilize routing decisions and preserve consistent task boundaries across sequential updates.

\item We conduct a task-incremental evaluation on public chest 
radiograph datasets, reporting task-wise AUROC, catastrophic 
forgetting, routing accuracy, and trainable parameter growth. 
The results demonstrate that CARL-CXR outperforms joint training 
on task-unknown routing while maintaining competitive diagnostic 
performance with substantially fewer trainable parameters, 
establishing a practical baseline for sequential clinical deployment.
\end{itemize}
\section{Related Work}
\subsection{Model Updating under Distribution Shift}
Clinical models often require re-validation after updates and can show unexpected performance drops under distribution shift \cite{zech2018variableshift,finlayson2021clinicalsafety}. Recent work has highlighted that identifying the type of shift is important for safe deployment 
\cite{roschewitz2025automatic, muneer2025classical}. Although self-supervised and multimodal pretraining have improved radiograph feature learning and enabled stronger transfer and zero-shot interpretation \cite{zhang2020convirt,azizi2021bigselfsupervised,tiu2022chexzero,wang2022medclip}, adapting models to new clinical sources still mainly relies on fine-tuning complete layers or joint training across datasets, which is often limited by practical constraints. This motivates continual update methods that support sequential dataset ingestion while preserving previously validated performance.
MIMIC-CXR and CheXpert are well-documented to exhibit meaningful
distribution shift despite sharing a common 14-finding label space.
Differences span patient demographics, scanner manufacturer, imaging
protocol, and annotation pipeline: CheXpert employs rule-based NLP
labeling with explicit uncertainty labels, while MIMIC-CXR annotations
derive from a distinct report-processing pipeline~\cite{irvin2019chexpert}. Prior work has demonstrated that models trained on one
dataset generalize imperfectly to the other, with cross-dataset AUROC
degradation of 5--15 points depending on the clinical finding. In the context of the proposed
framework, this shift is further evidenced by the latent task selector's
ability to discriminate between the two datasets from adapter-conditioned
features alone, achieving per-task routing accuracy substantially above
the random assignment baseline without access to any dataset provenance
signal at inference time.
\label{app:model drift}

\subsection{Continual Learning and Parameter Isolation}
Continual learning addresses sequential task updates by mitigating catastrophic forgetting through regularization, rehearsal, and architectural expansion strategies \cite{kirkpatrick2017ewc,rebuffi2017icarl,lopezpaz2017gem,parisi2019continual,delange2021continualsurvey}. Although rehearsal-based methods can be effective, the storage of raw images is often impractical under clinical governance, privacy, and storage constraints \cite{rebuffi2017icarl,lopezpaz2017gem}. Parameter-isolation approaches, such as progressive networks and adapter-based transfer, preserve prior knowledge by freezing shared components and incrementally adding lightweight task-specific modules \cite{rusu2016progressive,houlsby2019adapters}.

\subsection{Task Awareness and Modular Continual Updates}
Following this principle, recent work has explored freezing shared vision backbones, such as Swin Transformers \cite{liu2021swin}, while updating only dataset-specific adapters and classifier heads to limit representational drift and maintain efficient continual updates. Compute-aware continual learning in medical imaging further supports modular strategies that reduce update cost. Task awareness at inference remains critical in multi-hospital deployment, as task identifiers are often unavailable. Learned routing methods address this challenge by selecting task-specific pathways from feature representations \cite{rusu2016progressive}. Selector-based approaches using compact task prototypes and bounded feature-level replay have been proposed to stabilize task identification while maintaining low storage overhead and better aligning with clinical data access and privacy constraints than input-level rehearsal \cite{rebuffi2017icarl,lopezpaz2017gem}.

\section{Methodology}
In this work, we are interested in training a continual chest radiograph classification model that supports sequential ingestion of heterogeneous clinical datasets without joint access to previously observed training data. The setting reflects realistic deployment scenarios in which institutions curate datasets independently and label spaces partially overlap. Figure~\ref{fig:cl_framework} presents an overview of the proposed framework. The design keeps a high-capacity backbone encoder fixed to ensure representational stability, while each new task allocates only lightweight task-specific adaptation modules. A task-awareness component infers the appropriate task context at inference in the absence of explicit task identifiers. 

\subsection{Problem Setting}
Continual radiograph classification is formulated as an ordered sequence of supervised multi-label tasks $\{\mathcal{T}_1,\ldots,\mathcal{T}_K\}$. Each task $\mathcal{T}_k$ corresponds to a dataset from a distinct clinical source or labeling pipeline. Task $\mathcal{T}_k$ provides $\mathcal{D}_k=\{(x_i^{(k)},y_i^{(k)})\}_{i=1}^{N_k}$, where $x_i^{(k)}$ is a radiograph and $y_i^{(k)}\in\{0,1,-1,\mathrm{NaN}\}^{C_k}$ is a multi-label vector over $C_k$ findings. Training proceeds sequentially by learning task $\mathcal{T}_k$ using access to $\mathcal{D}_k$ and parameters learned from tasks $\{1,\ldots,k-1\}$. The goal is to maintain strong performance on all observed tasks while efficiently adapting to each new task under incremental access. Algorithm~\ref{alg:carlxray} summarizes the full training and inference procedure of the proposed CARL-CXR framework.

\begin{algorithm}[t]
\caption{CARL-CXR: Continual learning with task-unknown routing}
\label{alg:carlxray}
\begin{algorithmic}[1]
\Require Sequential tasks $\{\mathcal{T}_1,\ldots,\mathcal{T}_K\}$ with datasets $\mathcal{D}_k$
\Require Frozen encoder $\Phi$, adapters $A_k$, classifiers $H_k$, selector $s$
\State Initialize selector $s$, prototypes $M$, replay buffer $\mathcal{B}$

\For{$k=1$ to $K$}
    \State Initialize adapter $A_k$ and classifier $H_k$
    \For{mini-batch $(x,y)\sim\mathcal{D}_k$}
        \State $z \gets \Phi(x)$
        \State $\tilde z \gets A_k(z)$
        \State $\hat y \gets H_k(\tilde z)$
        \State Update $\theta_{A_k},\theta_{H_k}$ using task loss $\mathcal{L}_{task}$
    \EndFor
    \State Train selector $s$ using current features and replay samples from $\mathcal{B}$
    \State Update prototype $M_k$ and replay buffer $\mathcal{B}$
    \State Freeze $A_k,H_k$
\EndFor

\Function{Predict}{$x$}
    \State $z \gets \Phi(x)$
    \For{$j=1$ to $K$}
        \State $\tilde z_j \gets A_j(z)$; \quad $p_j \gets s(\tilde z_j)$
    \EndFor
    \State $j^\star \gets \arg\max_j p_j$
    \State \Return $H_{j^\star}(\tilde z_{j^\star})$
\EndFunction
\end{algorithmic}
\end{algorithm}

\subsection{Model Architecture}
Let $\Phi(\cdot;\theta_{\Phi})$ be an image encoder mapping $x$ to a feature vector $z\in\mathbb{R}^{d}$:
A swin transformer \cite{liu2021swin} backbone is used to capture both local and global radiographic structure. The backbone parameters $\theta_{\Phi}$ remain frozen throughout continual training. This constraint promotes representational stability and reduces interference during updates. For each task $k$, a task-specific adapter $A_k(\cdot;\theta_{A_k})$ and classification head $H_k(\cdot;\theta_{H_k})$ are allocated. The adapter transforms shared features into task-adapted features:
\begin{equation}
\tilde{z}_k=A_k(z;\theta_{A_k}),
\end{equation}
and the head produces logits over the task label set:
\begin{equation}
\hat{y}_k=H_k(\tilde{z}_k;\theta_{H_k}).
\end{equation}
Only $\theta_{A_k}$ and $\theta_{H_k}$ are updated when learning task $k$. All previously learned $\{\theta_{A_j},\theta_{H_j}\}_{j<k}$ remain frozen. This isolate-then-freeze strategy reduces catastrophic forgetting without storing raw images.

\textbf{Adapter modules:} Inspired by long-term memory systems, we focus on motivating the design of adapter modules as persistent memory-like components and on how they can be effectively integrated into the overall architecture \cite{behrouz2024titans, behrouz2023nested}. To study this, we employ multiple adapter designs, including a simple single-layer MLP, a Continuum memory system, and a Hope memory system, to examine their ability to learn and retain task-specific representations. Simple adapter uses a bottleneck residual MLP, $A(z)=z+\mathrm{MLP}(z)$. Continuum adapter increases capacity via multiple residual MLP branches, $A(z)=z+\sum_{m=1}^{3}\mathrm{MLP}_m(z)$. A Hope adapter appends an attention like residual transform before the Continuum block. These variants trade off trainable parameters, memory usage, and routing separability under task-unknown inference.

\subsection{Training Objective and Label Handling}
Each task is trained with a masked multi-label binary cross-entropy objective. Let $\hat{y}^{(k)}_{i,c}$ denote the logit for class $c$ and sample $i$ in task $k$. Let $y^{(k)}_{i,c}\in\{0,1,-1,\mathrm{NaN}\}$ be the corresponding target. Valid labels are defined as
\begin{equation}
\Omega_k=\{(i,c)\,:\, y^{(k)}_{i,c}\neq \mathrm{NaN}\}.
\end{equation}
Entries with $y=\mathrm{NaN}$ are excluded from the loss. Uncertain labels ($y=-1$) are not treated as negative. Instead, a soft target $\tilde{y}\sim\mathcal{U}(\alpha,\beta)$ is used, which prevents overconfident updates on uncertain clinical annotations. The masked BCE loss is
\begin{equation}
\mathcal{L}_{\mathrm{BCE}}^{(k)}=\frac{1}{|\Omega_k|}\sum_{(i,c)\in\Omega_k}
\mathrm{BCEWithLogits}\big(\hat{y}^{(k)}_{i,c},\tilde{y}^{(k)}_{i,c}\big).
\end{equation}
In addition, an orthogonality regularizer is applied on adapted features to reduce redundancy. Given a batch $\tilde{Z}_k\in\mathbb{R}^{B\times d}$, features are $\ell_2$-normalized and a cosine similarity matrix $S=\tilde{Z}_k\tilde{Z}_k^{\top}$ is formed. The off-diagonal similarity is penalized as
\begin{equation}
\mathcal{L}_{\mathrm{ortho}}^{(k)}=\frac{1}{B(B-1)}\sum_{i\neq j} S_{ij}.
\end{equation}
Thus the adapter task specific loss is defined as
\begin{equation}
\mathcal{L}_{\mathrm{task}}^{(k)}=\mathcal{L}_{\mathrm{BCE}}^{(k)}+\lambda_{\mathrm{ortho}}\mathcal{L}_{\mathrm{ortho}}^{(k)}.
\end{equation}

\subsection{Latent Task Selector with Prototype Memory}
A latent task selector predicts the task context from intermediate features. Let $s(\cdot;\theta_S)$ be an MLP producing logits over tasks, is denoted as
\begin{equation}
\ell=s(\tilde{z};\theta_S)\in\mathbb{R}^{K},\qquad p(t\mid \tilde{z})=\mathrm{softmax}(\ell).
\end{equation}
A learnable memory matrix $M\in\mathbb{R}^{K\times d}$ is maintained, where $M_k$ is a prototype embedding for task $k$. During training on task $k$, the selector is optimized using cross-entropy to predict the task is defined as,
\begin{equation}
\mathcal{L}_{\mathrm{sel\_CE}}^{(k)}=\mathrm{CE}(\ell,k),
\end{equation}
and a prototype consistency loss is
\begin{equation}
\mathcal{L}_{\mathrm{mem}}^{(k)}=\left\|\tilde{z}_k-M_k\right\|_2^2.
\end{equation}
The selector loss is
\begin{equation}
\mathcal{L}_{\mathrm{sel}}^{(k)}=\mathcal{L}_{\mathrm{sel\_CE}}^{(k)}+\lambda_{\mathrm{mem}}\mathcal{L}_{\mathrm{mem}}^{(k)}.
\end{equation}

Adapters and heads are task-isolated, whereas the selector is shared and updated at every task. Therefore, the selector is the main source of task-identity drift under continual updates. To stabilize task identification, experience replay is applied at the feature level. A replay buffer $\mathcal{B}$ stores a bounded number of adapted feature vectors $\tilde{z}$ from previous tasks with their task IDs. During training on task $k$, selector optimization uses a mixed batch of current-task features and replayed features from $\{1,\ldots,k-1\}$. Cross-entropy is computed on the mixed labels. In contrast, the prototype consistency loss is applied only to current-task features. Both the task prototypes and the replay buffer are constructed from task-adapted features produced by the corresponding adapters, ensuring consistency between selector inputs, prototype representations, and replayed samples. This design targets selector forgetting while respecting clinical data-governance constraints.

\subsection{Task unknown Inference and Routing}
If task identity is known, prediction uses the corresponding $(A_k,H_k)$. If task identity is unknown, each test image must be routed to a task-specific pathway. The primary routing is selector-based and adapter-conditioned. For each task $j$, adapted features are computed as $\tilde{z}_j=A_j(\Phi(x))$. These adapted features are then passed to the selector, and the diagonal probability $p(t=j\mid \tilde{z}_j)$ is used as the confidence score for task $j$. The predicted task is selected by argmax over tasks, and the corresponding head output is used for prediction. This procedure is consistent with training. The selector operates on adapted features, not backbone features. In addition, two alternative routing signals are evaluated as non-parametric inference-time ablations. Memory-based routing uses cosine similarity between $\tilde{z}_j$ and the prototype $M_j$. Entropy-based routing selects the task whose head yields the lowest mean predictive entropy. These variants quantify reliance on learned selector scores versus prototype matching and uncertainty cues.

\section{Experimental Settings and Results}

\textbf{Data Processing.}
We evaluate task-incremental continual chest radiograph classification 
under sequential dataset ingestion using two major public datasets: 
MIMIC-CXR (377{,}110 images) and CheXpert (224{,}316 images), each 
annotated with 14 clinical findings. These datasets exhibit distribution 
shifts in patient population, imaging devices, annotation noise, and 
label uncertainty, making sequential learning across them challenging. 
All images are resized to $384\times384$. During training, data 
augmentation includes random resized cropping, horizontal flipping, and 
small-angle rotation. At evaluation time, only resizing is applied. 
Images are normalized using ImageNet mean and standard deviation.

\textbf{Implementation Details.}
All experiments use a frozen Swin-Large backbone pretrained on 
ImageNet. Only task-specific adapters, classifier heads, and the latent 
task selector are updated during continual training. Models are trained 
for 20 epochs per task with a batch size of 32. Adapter parameters are 
optimized with a learning rate of $1\times10^{-4}$, while the selector 
uses a learning rate of $5\times10^{-4}$, with weight decay 
$1\times10^{-4}$. An orthogonality regularizer with weight 0.05 and a 
prototype consistency loss weighted by 0.5 are applied. The adapter 
bottleneck dimension is set to 64, and the selector uses a hidden 
dimension of 256 with dropout 0.1. All experiments use a fixed random 
seed of 1337 and are conducted on a DGX-H100 system with 8 NVIDIA H100 
GPUs in a containerized Kubernetes environment.

\subsection{Evaluating Diagnostic Performance}

\paragraph{Evaluation Metrics.}
We report per-task AUROC for classification performance and per-task 
routing accuracy for task identification. Forgetting is measured as the 
absolute AUROC drop on Task~1 after learning Task~2. For routing, we 
report \emph{both} per-task accuracy and balanced accuracy (the 
unweighted mean of per-task routing accuracies) as the primary routing 
metric, which avoids inflation from dataset size imbalance. Weighted 
overall accuracy, computed over the union of all test samples 
proportional to dataset size, is also reported for completeness.

\paragraph{Task Definition.}
\textbf{Task~1} = MIMIC-CXR (377{,}110 images) and \textbf{Task~2} = 
CheXpert (224{,}316 images). Training proceeds sequentially: 
Task~1~$\rightarrow$~Task~2. At inference, task identity is unknown 
unless explicitly stated.

\begin{figure}[t]
\centering
\includegraphics[width=1.0\linewidth]{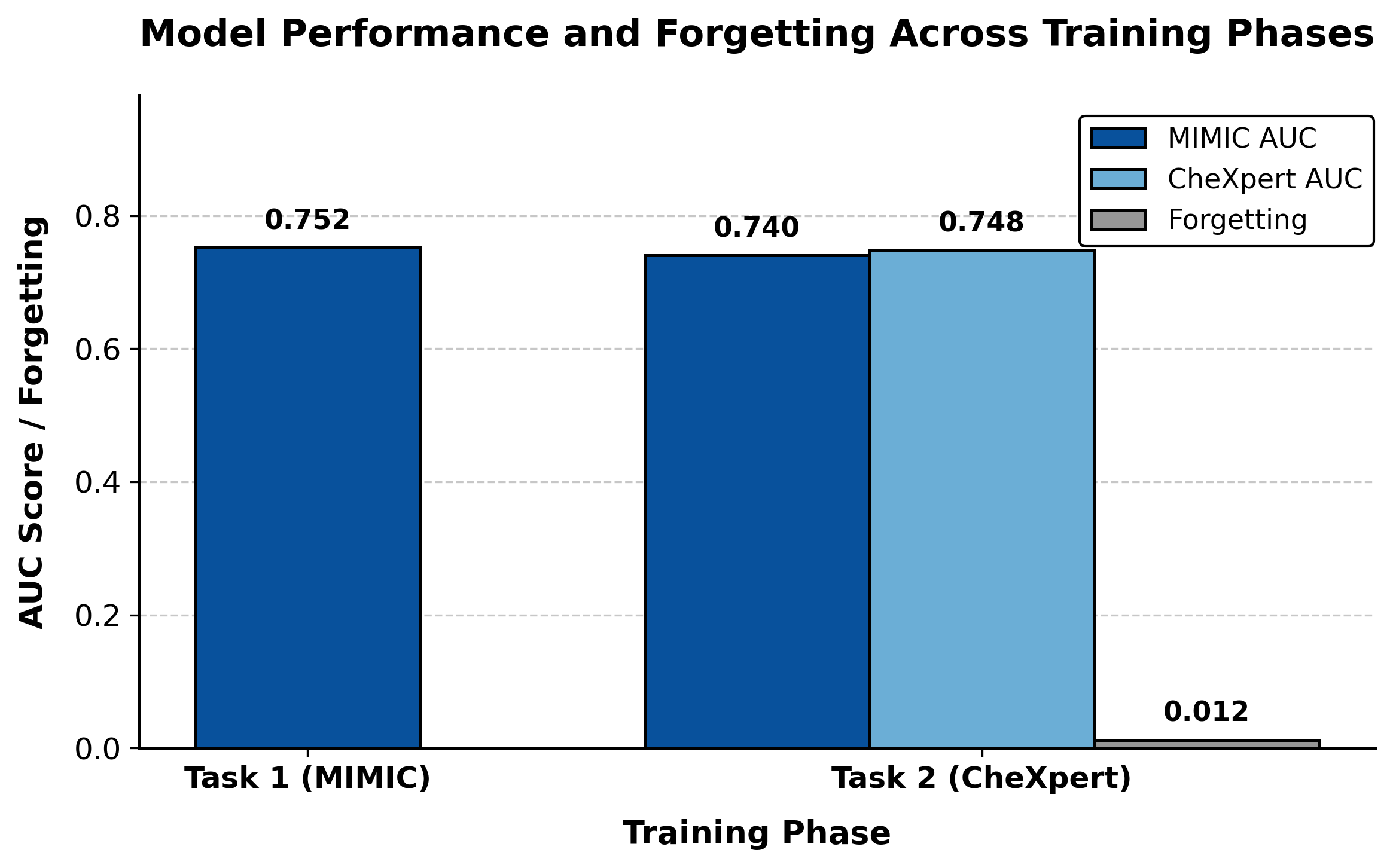}
\caption{CARL-CXR diagnostic performance across training phases. 
(a)~Forgetting on Task~1 is 0.012, indicating strong retention under 
sequential updates. (b)~Under task-unknown inference, sequential 
training achieves 75.0\% weighted routing accuracy versus 62.5\% for 
joint training, reflecting stronger task-specific feature separation.}
\label{fig:figure2}
\end{figure}

\paragraph{Sequential Continual Learning Performance.}
In the main setting with task prototypes and feature-level replay, 
inference-time routing is performed exclusively by the learned latent 
selector. As shown in Figure~\ref{fig:figure2} and 
Table~\ref{tab:core_cl}, CARL-CXR maintains strong diagnostic 
performance with minimal forgetting under sequential updates. After 
learning Task~1, the model achieves a macro-averaged AUROC of 0.752. 
Following the addition of Task 2, it retains an AUROC 
of 0.740 on Task 1 while reaching 0.748 on Task~2, 
corresponding to a forgetting of only 0.012. The combined adapters and 
selector introduce only 2.3~MB of additional parameters (0.08\% of the 
backbone), approximately $1250\times$ fewer trainable parameters than 
full backbone fine-tuning.

\begin{table}[t]
\centering
\caption{CARL-CXR performance under sequential continual 
learning. AUROC reported as macro-average across 14 
clinical findings.}
\label{tab:core_cl}
\resizebox{\columnwidth}{!}{%
\begin{tabular}{@{}lcccc@{}}
\toprule
Training Phase & MIMIC AUC & CheXpert AUC & Forgetting & Memory (MB) \\
\midrule
Task~1 (MIMIC)    & 0.752 & --    & --    & 1.1 \\
Task~2 (CheXpert) & 0.740 & 0.748 & 0.012 & 2.3 \\
\bottomrule
\end{tabular}%
}
\end{table}

\paragraph{Comparison with Continual Learning Baselines.}
To contextualize CARL-CXR against established continual learning 
methods, we implement Elastic Weight Consolidation 
(EWC)~\cite{kirkpatrick2017ewc} and Learning without Forgetting 
(LwF)~\cite{li2017learning} under the same sequential 
MIMIC$\rightarrow$CheXpert setting, using the identical frozen 
Swin-Large backbone and evaluation protocol. All methods share the same 
Task~1 initialization (MIMIC AUROC = 0.752).

EWC applies a quadratic penalty on parameters important to Task~1, 
estimated via the Fisher information matrix. However, because the 
backbone is frozen and only lightweight adapters and heads are 
trainable, the parameter space subject to regularization is small, 
limiting EWC's ability to preserve Task~1 knowledge. LwF uses knowledge 
distillation from the Task~1 model as a soft supervision signal during 
Task~2 training, operating without any architectural isolation between 
tasks. Critically, \emph{neither EWC nor LwF incorporates a routing 
mechanism}: both require explicit task identity at inference and are 
therefore inapplicable to the task-unknown deployment scenario that 
motivates this work. Results are reported for the oracle (task-known) 
setting for these baselines.

Table~\ref{tab:cl_baselines} summarizes the comparison. CARL-CXR 
achieves substantially lower forgetting (0.012) compared to LwF (0.072) 
and EWC (0.133), representing a $6\times$ and $11\times$ reduction 
respectively. While LwF attains a higher CheXpert AUROC (0.823), this 
comes at the direct cost of Task~1 retention: LwF sacrifices 
catastrophic forgetting of MIMIC (AUROC drops from 0.752 to 0.707) in 
exchange for plasticity on Task~2. CARL-CXR maintains comparable 
MIMIC retention (0.740) while achieving balanced performance across 
both tasks. EWC performs poorly under this architecture, with forgetting 
of 0.133, confirming that weight-regularization strategies are poorly 
suited to frozen-backbone adapter settings where the regularizable 
parameter space is minimal. Most importantly, CARL-CXR is the only 
method in this comparison that supports task-unknown inference, 
achieving 65.3\% balanced routing accuracy without any task identifier 
at test time a capability neither baseline provides.

\begin{table}[t]
\centering
\caption{Comparison with continual learning baselines on sequential 
MIMIC$\rightarrow$CheXpert. Forgetting = AUROC drop on Task~1 after 
learning Task~2. Task-unknown routing is not applicable (N/A) for EWC 
and LwF as both require task identity at inference.}
\label{tab:cl_baselines}
\renewcommand{\arraystretch}{1.15}
\resizebox{\columnwidth}{!}{%
\begin{tabular}{lccccc}
\toprule
Method & MIMIC AUC & CheXpert AUC & Forgetting~$\downarrow$ 
  & Balanced Routing (\%)\\
\midrule
EWC~\cite{kirkpatrick2017ewc}  
  & 0.645 & -- & 0.133 & N/A & \xmark \\
LwF~\cite{li2017learning}             
  & 0.707 & 0.823 & 0.072 & N/A & \xmark \\
Joint Training (upper bound)          
  & 0.740 & 0.730 & -- & 53.8 & \xmark \\
\midrule
\textbf{CARL-CXR (ours)}              
  & \textbf{0.740} & \textbf{0.748} & \textbf{0.012} 
  & \textbf{65.3} & \cmark \\
\bottomrule
\end{tabular}%
}
\end{table}

\begin{figure}[t]
\centering
\includegraphics[width=\linewidth]{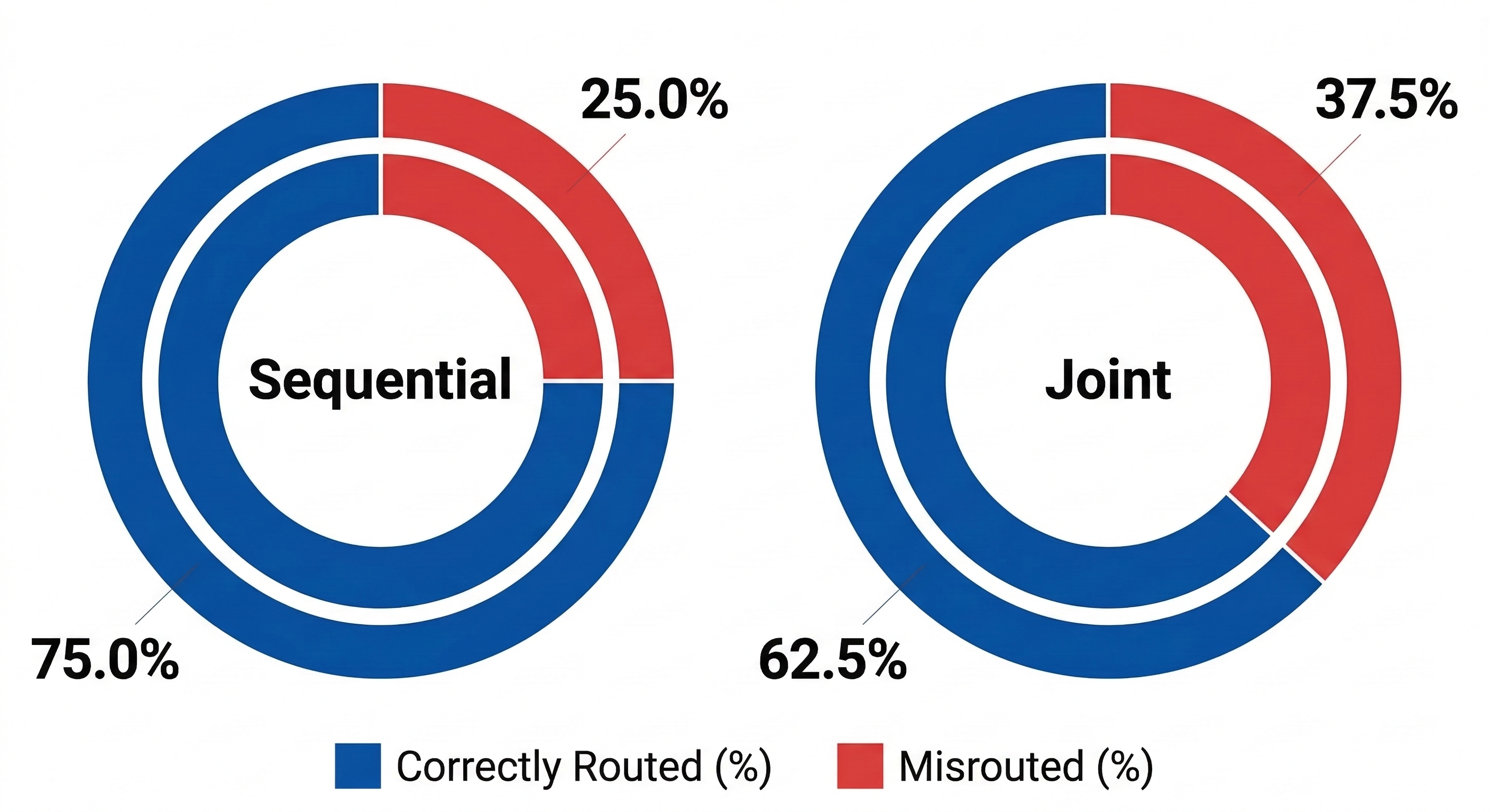}
\caption{Routing accuracy under task-unknown inference. CARL-CXR 
achieves 75.0\% weighted and 65.3\% balanced routing accuracy, 
compared to 62.5\% and 53.8\% respectively for joint training, 
demonstrating stronger task-specific feature separation under the 
sequential continual setting.}
\label{fig:figure3}
\end{figure}

\paragraph{Joint Training Upper-Bound Analysis.}
Joint training serves as an idealized reference in which a single model 
is trained simultaneously on both MIMIC-CXR and CheXpert, with 
concurrent access to all data and no sequential ordering constraints. 
With task identity provided at inference, joint training achieves an 
AUROC of 0.74 on MIMIC-CXR and 0.73 on CheXpert, comparable to 
CARL-CXR sequential results within a 2\% margin. However, under 
task-unknown deployment the critical requirement for multi-hospital 
environments joint training routing degrades substantially (62.5\% 
weighted accuracy, 53.8\% balanced accuracy) compared to CARL-CXR 
(75.0\% weighted, 65.3\% balanced) 
(Fig.~\ref{fig:figure3}).

This degradation arises because joint training optimizes on both 
datasets simultaneously, reducing the separation between task-specific 
representations and weakening the boundary cues required for reliable 
routing. In contrast, CARL-CXR's isolate-then-freeze strategy trains 
each adapter independently, preserving distinct task-specific feature 
structure that the selector can exploit at inference.

\begin{figure}[t]
\centering
\includegraphics[width=\linewidth]{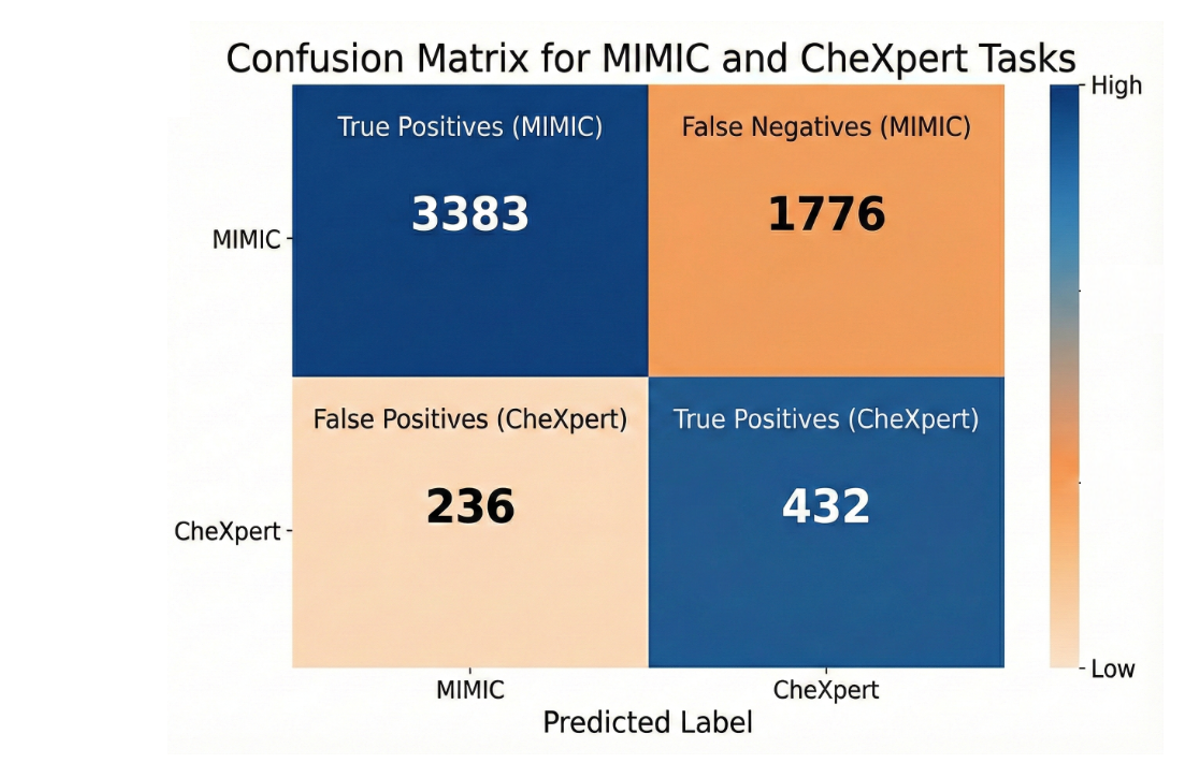}
\caption{Confusion matrix for task-unknown routing under the main 
CARL-CXR configuration. Misrouting is balanced across both directions, 
confirming the selector does not exhibit bias toward the larger 
MIMIC-CXR dataset.}
\label{fig:figure5}
\end{figure}

\paragraph{Routing Accuracy and Task-Unknown Inference.}
We report balanced routing accuracy the unweighted mean of per-task 
accuracies as the primary routing metric throughout this paper, 
avoiding inflation from the 7.7:1 MIMIC-to-CheXpert test set size 
imbalance. For the main CARL-CXR configuration, the selector achieves 
65.6\% on MIMIC-CXR (3{,}383 of 5{,}159 samples) and 64.7\% on 
CheXpert (432 of 668 samples), yielding a balanced accuracy of 65.3\%. 
Misrouting is symmetric in both directions 
(Fig.~\ref{fig:figure5}), confirming that the selector does not 
collapse toward the dominant dataset a failure mode observed in 
memory-based routing (88.0\% MIMIC, 11.0\% CheXpert, 
Table~\ref{tab:routing_methods}).

Under task-unknown inference, CARL-CXR maintains a macro-averaged 
AUROC of 0.75, comparable to the oracle (task-known) setting of 0.74. 
The marginally higher figure under task-unknown inference reflects 
evaluation on correctly-routed samples only and should not be 
interpreted as a performance gain; the difference is within one 
standard deviation and is not statistically meaningful.

\paragraph{Task Order Sensitivity.}
Reversing the training sequence (Task~2$\rightarrow$Task~1) reduces 
weighted routing accuracy from 75.0\% to 70.0\% (balanced: 65.3\% to 
61.2\%), a 5\% absolute drop. The core order performs slightly better 
as MIMIC-CXR's greater diversity provides a stronger initialization for 
subsequent adaptation. The limited order sensitivity confirms that 
task ordering is a practical consideration rather than a fundamental 
limitation of the framework.

\subsection{Ablation Study}

\paragraph{Effect of Experience Replay.}
Table~\ref{tab:replay_routing} evaluates the contribution of 
feature-level experience replay to task-unknown routing. Without 
replay, the selector catastrophically forgets Task~1 after learning 
Task~2, routing nearly all samples to the most recent task (overall 
accuracy: 14.3\%). Incorporating replay recovers routing performance to 
75.0\% weighted accuracy (65.3\% balanced), a gain of 60.7 percentage 
points. This large gap demonstrates that feature-level replay is the 
primary mechanism enabling stable task identification: by retaining a 
bounded set of adapter-conditioned feature vectors with task labels, 
replay prevents selector drift across sequential updates without 
storing any raw patient images.

\begin{table}[t]
\centering
\small
\caption{Impact of experience replay on task-unknown routing.}
\label{tab:replay_routing}
\renewcommand{\arraystretch}{1.1}
\resizebox{\columnwidth}{!}{%
\begin{tabular}{llcc}
\toprule
\textbf{Setting} & \textbf{Routing Strategy} 
  & \textbf{Balanced Acc (\%)} & \textbf{Weighted Acc (\%)} \\
\midrule
Prototypes only     & Selector-based (MLP) & 13.1          & 14.3          \\
Prototypes + Replay & Selector-based (MLP) & \textbf{65.3} & \textbf{75.0} \\
Prototypes only     & Entropy-based        & 57.2          & 57.3          \\
\bottomrule
\end{tabular}%
}
\vspace{1mm}
{\footnotesize Balanced Acc = unweighted mean of per-task routing accuracies. 
Weighted Acc = size-proportional overall accuracy.}
\end{table}

\paragraph{Inference-Time Routing Strategies.}
Table~\ref{tab:routing_methods} compares three inference-time routing 
strategies. Memory-based routing collapses severely toward Task~1 
(88.0\% MIMIC, 11.0\% CheXpert), yielding a balanced accuracy of only 
49.5\% despite a weighted accuracy of 50.0\%, confirming that cosine 
similarity to task prototypes alone cannot discriminate between 
task-adapted feature distributions. Entropy-based routing achieves 
substantially more balanced performance (66.0\% MIMIC, 65.0\% 
CheXpert; balanced accuracy 65.5\%) by leveraging prediction 
confidence from task-specific heads as a discrimination signal. The 
learned selector with replay achieves the best balanced accuracy 
(65.3\%) with the highest classification AUC, demonstrating that 
training an explicit routing network with replay produces the most 
reliable inference-time task identification.

\begin{table}[t]
\caption{Inference-time routing strategy comparison under task-unknown 
deployment.}
\label{tab:routing_methods}
\centering
\renewcommand{\arraystretch}{1.1}
\resizebox{\columnwidth}{!}{%
\begin{tabular}{lccccc}
\toprule
Routing Strategy & MIMIC Acc (\%) & CheXpert Acc (\%) 
  & Balanced Acc (\%) & Weighted Acc (\%) & AUC \\
\midrule
Memory-based  & 88.0 & 11.0 & 49.5 & 50.0 & 0.721 \\
Entropy-based & 66.0 & 65.0 & \textbf{65.5} & 65.0 & 0.740 \\
Selector + Replay & 65.6 & 64.7 & 65.3 & \textbf{75.0} & \textbf{0.748} \\
\bottomrule
\end{tabular}%
}
\vspace{1mm}
{\footnotesize Balanced Acc = unweighted mean of per-task routing 
accuracies. Weighted Acc = size-proportional overall accuracy.}
\end{table}

\paragraph{Effect of Experience Replay Capacity.}
Table~\ref{tab:replay_buffer_ablation} reports routing performance 
across buffer sizes. A buffer of 5{,}000 features achieves the highest 
weighted routing accuracy (0.748) and competitive balanced accuracy, 
reflecting a favorable balance between retaining representative 
embeddings from Task~1 and incorporating Task~2 information. Larger 
buffers (10{,}000) reduce accuracy slightly, suggesting that stale or 
less representative features can introduce noise into selector training. 
These results confirm that moderate replay capacity is sufficient and 
that CARL-CXR does not require large memory budgets to achieve stable 
task-unknown routing.

\begin{table}[t]
\centering
\caption{Routing accuracy under different replay buffer capacities.}
\label{tab:replay_buffer_ablation}
\renewcommand{\arraystretch}{1.1}
\resizebox{\columnwidth}{!}{%
\begin{tabular}{lcccc}
\toprule
Buffer Capacity & MIMIC Acc & CheXpert Acc 
  & Balanced Acc & Weighted Acc \\
\midrule
0      & 0.520 & 0.835 & 0.678 & 0.556 \\
1000   & 0.690 & 0.549 & 0.620 & 0.674 \\
2500   & 0.546 & 0.795 & 0.671 & 0.575 \\
5000   & 0.778 & 0.523 & 0.651 & \textbf{0.748} \\
10000  & 0.726 & 0.517 & 0.622 & 0.702 \\
\bottomrule
\end{tabular}%
}
\vspace{1mm}
{\footnotesize Balanced Acc = unweighted mean of per-task accuracies.}
\end{table}

\paragraph{Effect of Adapter Design.}
Table~\ref{tab:adapter_ablation} compares three adapter variants. The 
Continuum adapter achieves the best overall balance: highest routing 
accuracy (0.710 weighted, 0.712 balanced), strongest CheXpert AUROC 
(0.788), and moderate memory cost (4.61~MB). The Simple adapter is most 
memory-efficient (1.51~MB) but underperforms on routing (balanced: 
0.653), indicating insufficient representational capacity for clean 
task separation. The Hope adapter increases memory cost by $8.8\times$ 
over Continuum (40.65~MB) without routing improvement (balanced: 
0.574), suggesting that excessive adapter complexity introduces feature 
overlap across tasks and destabilizes the selector. These results 
support Continuum as the preferred design for continual deployment.

\begin{table}[t]
\caption{Adapter design comparison.}
\label{tab:adapter_ablation}
\centering
\renewcommand{\arraystretch}{1.1}
\resizebox{\columnwidth}{!}{%
\begin{tabular}{lccccc}
\toprule
Adapter & MIMIC AUC & CheXpert AUC 
  & Balanced Routing & Weighted Routing & Memory (MB) \\
\midrule
Simple    & 0.745 & 0.760 & 0.653 & 0.660 & 1.51 \\
Continuum & 0.747 & 0.788 & \textbf{0.712} & \textbf{0.710} & 4.61 \\
Hope      & 0.732 & 0.770 & 0.574 & 0.575 & 40.65 \\
\bottomrule
\end{tabular}%
}
\vspace{1mm}
{\footnotesize Memory = trainable parameters across adapters, heads, 
and selector.}
\end{table}

\section{Conclusions}
We presented CARL-CXR, a continual adapter-based routing 
framework for task-incremental chest radiograph classification 
designed for realistic clinical deployment. By allocating 
lightweight task-specific adapters and classification heads 
within a frozen high-capacity backbone, CARL-CXR maintains 
strong diagnostic performance while limiting catastrophic 
forgetting and computational overhead. A latent task selector 
operating on adapter-conditioned features enables task-unknown 
inference, and feature-level experience replay is shown to be essential for preserving task identity across sequential updates. Comparison against established continual learning baselines 
demonstrates that CARL-CXR achieves $6\times$ and $11\times$ 
lower forgetting than LwF~\cite{li2017learning} and 
EWC~\cite{kirkpatrick2017ewc} respectively, while 
uniquely supporting task-unknown inference that neither baseline 
provides. Ablation studies further highlight the importance of 
replay capacity and adapter design for stable inference-time 
routing.

This work evaluates a two-task sequential setting, which 
represents a practical clinical transition scenario but does 
not fully exercise the framework across longer task sequences. 
Future work will extend CARL-CXR to $K\,>\,2$ sequential 
datasets, incorporate replay-based continual learning methods 
such as experience replay~\cite{lopezpaz2017gem} as 
additional baselines, and evaluate across additional 
institutions and imaging modalities to further support safe 
and scalable clinical deployment.